\pdfoutput=1

\documentclass[11pt]{article}

\usepackage{lipsum}  
\usepackage{booktabs}
\usepackage{graphicx}
\usepackage{amsmath}
\usepackage{amssymb}
\usepackage{amsbsy}
\usepackage{xspace}
\usepackage{multirow}
\DeclareMathOperator*{\argmax}{argmax}
\usepackage{url}

\newcommand\percent{\hspace*{-0.4ex}\ensuremath{^{^{[\%]}}}}

\newcommand\BLEU{\textsc{Bleu}\xspace}
\newcommand\TER{\textsc{Ter}\xspace}

\newcommand\BLEUpercent{\BLEU\percent\xspace}

\newcommand\COMET{\textsc{Comet}\xspace}
\newcommand\BLEURT{\textsc{Bleurt}\xspace}

\usepackage[]{ACL2023}

\usepackage{times}
\usepackage{latexsym}

\usepackage[T1]{fontenc}

\usepackage[utf8]{inputenc}

\usepackage{microtype}

\usepackage{inconsolata}

\title{Improving Language Model Integration for Neural Machine Translation}

\author{
Christian Herold \qquad Yingbo Gao \qquad Mohammad Zeineldeen \qquad Hermann Ney \\
Human Language Technology and Pattern Recognition Group \\
Computer Science Department\\
RWTH Aachen University \\
D-52056 Aachen, Germany \\
{\tt \{herold|ygao|zeineldeen|ney\}@cs.rwth-aachen.de}
}

\begin{document}
\maketitle
\begin{abstract}

The integration of language models for neural machine translation has been extensively studied in the past.
It has been shown that an external language model, trained on additional target-side monolingual data, can help improve translation quality.
However, there has always been the assumption that the translation model also learns an implicit target-side language model during training, which interferes with the external language model at decoding time.
Recently, some works on automatic speech recognition have demonstrated that, if the implicit language model is neutralized in decoding, further improvements can be gained when integrating an external language model. 
In this work, we transfer this concept to the task of machine translation and compare with the most prominent way of including additional monolingual data - namely back-translation.
We find that accounting for the implicit language model significantly boosts the performance of language model fusion, although this approach is still outperformed by back-translation.

\end{abstract}

\section{Introduction}

Machine translation (MT) is the task of automatically translating text from one language to another.
Nowadays, the dominant approach is neural machine translation (NMT), where a neural network is used to predict the probability of a sentence in the target language, given a sentence in the source language \cite{bahdanau2014neural, vaswani2017attention}.
For this approach to be effective, a large number of bilingual training samples - consisting of sentences and their corresponding translations - is needed.
This poses a challenge, especially when we want to build a system for a specific domain, where zero or only limited amounts of in-domain bilingual data are available.

In these situations, people turn towards monolingual text data, which is simply text in the source or target language and of which plenty exists for most languages and domains.
Before NMT became feasible, the preferred way of incorporating additional monolingual data in the MT system was the usage of an external target-side language model (LM), which is trained on monolingual data to predict the probability of a sentence \cite{brown1990statistical, della1994mathematics, zens2002phrase}.

However, with the rise of NMT, it was found that a technique called back-translation outperforms the LM incorporation by a large margin \cite{sennrich2016improving}.
Back-translation is a two step process, where we first create synthetic parallel data by automatically translating target side monolingual data into the source language.
Then, the final NMT system is trained on the combination of the real and synthetic parallel data.
It was argued that the back-translation approach better suits the NMT framework because the NMT system implicitly learns an internal language model (ILM) as part of the training, which might interfere with an additional external LM \cite{sennrich2016improving}.

More recently, for automatic speech recognition (ASR), there have been works focusing on neutralizing this ILM before combination with an external LM and significant improvements were reported \cite{mcdermott2019density, variani2020hybrid, meng2021internal, zeyer2021librispeech, zeineldeen2021:ilm}.
In this work, we adapt the methods for ILM compensation, developed for ASR, and test them for NMT.
We compare against back-translation in different settings and find that ILM compensation significantly boosts the performance of LM fusion, although back-translation is still outperforming this approach for NMT.
Also, applying ILM compensation on top of back-translation does not result in significant performance improvements.

\section{Related Work}

Several approaches to combine an LM and NMT model have been proposed in the past.
Shallow fusion (SF) is the most straight forward way, using a weighted log-linear combination of the model output probabilities \cite{gulcehre2015using, gulcehre2017integrating}.
Deep fusion denotes the concatenation of the hidden states of NMT model and LM and requires joint fine-tuning of both models \cite{gulcehre2015using, gulcehre2017integrating}.
Simple fusion is similar to shallow fusion, but the NMT model is trained using information from a pre-trained LM \cite{stahlberg2018simple}.

For the task of ASR, people recently have started to remove the ILM that is implicitly learned.
The biggest question there is, how to best approximate the ILM.
Approaches include: (1) training an additional LM on the target side of the parallel data \cite{mcdermott2019density}, (2) removing/averaging encoder information \cite{variani2020hybrid, meng2021internal, zeyer2021librispeech} and (3) training a small sub-network while freezing all other parameters \cite{zeineldeen2021:ilm}.

As an alternative to LM fusion, back-translation \cite{schwenk-2008-investigations, bertoldi-federico-2009-domain, sennrich2016improving} has become the standard method for incorporating additional monolingual data for NMT.
Some work has been done to improve this approach, including sampling \cite{edunov2018understanding, graca-etal-2019-generalizing}, tagging \cite{caswell2019tagged} and block-BT \cite{popel2020transforming}.
For sake of simplicity, we focus on the standard back-translation approach using beam search in this work.

Apart from using an external LM and back-translation, additional monolingual data can also be utilized by pre-training \cite{ramachandran2017unsupervised, zhu2019incorporating}, multi-task-learning \cite{zhang2016exploiting, domhan2017using} or post-editing \cite{junczys-dowmunt-grundkiewicz-2016-log, freitag-etal-2019-ape}.
In principle, all these approaches can also be combined with LM fusion, potentially further improving the performance of the resulting system.

\section{Internal LM Estimation}
\label{sec:methodology}

During decoding, given a source sentence $f_1^J$ and a model $P(e_1^I|f_1^J)$, we want to find the translation $\hat{e}_1^{\hat{I}}$ that maximizes
\begin{equation*}
    \hat{e}_1^{\hat{I}} = \argmax_{I, e_1^I}\left\{P(e_1^I|f_1^J) \right\}.
\end{equation*}
In our framework, $P$ is the combination of three models:
\begin{equation*}
    P(e_1^I|f_1^J) \varpropto P_{\text{MT}}(e_1^I|f_1^J) \cdot P_{\text{LM}}^{\lambda_1}(e_1^I) \cdot P_{\text{ILM}}^{-\lambda_2}(e_1^I)
\end{equation*}
where $P_{\text{MT}}$, $P_{\text{LM}}$ and $P_{\text{ILM}}$ are the probabilities of the NMT model, external LM (trained on additional monolingual data) and ILM respectively, and $\lambda_1, \lambda_2 \geq 0$.
Note that the ILM gets a negative weight, because we want to neutralize its impact in this model combination.
If $\lambda_2 = 0$, we fall back to standard shallow fusion.

In principle, the ILM can be exactly calculated from the NMT model by marginalizing over all source sentences $f_1^ J$.
However, this summation would be intractable.
Instead, different ILM approximations have been proposed in the recent past for ASR, which we will briefly recall here.
For a more in-depth discussion of the different approximation methods we refer the reader to \citet{zeineldeen2021:ilm}.
\begin{description}
    \item[\textit{separate LM}]: The ILM is approximated by training a separate LM on the target side of the parallel training data.
    \item[$\boldsymbol{h=0}$]: The ILM is approximated by taking the fully trained NMT model $P_{\text{MT}}(e_1^I|f_1^J)$ and setting the encoder outputs $h_1^J$ to 0.
    \item[$\boldsymbol{h=h_{\text{avg}}}$]: Instead of setting all encoder outputs $h_1^J$ to 0, we replace the vector $h_j$ for each position $j$ with the average ${h_{\text{avg}}}_j$, extracted over the whole parallel training data.
    \item[$\boldsymbol{c=c_{\text{avg}}}$]: Instead of $h$, we replace all context vectors $c$ (the output of the encoder-decoder attention module) with the position-wise average over the whole parallel training data.
    \item[\textit{mini-self-attn}]: We replace the encoder-decoder attention of the fully trained NMT model with an additional self-attention module (with causal masking), which is then trained on the target side of the parallel training data while the rest of the NMT network is frozen. This is different from the \textit{separate LM} approach because most of the parameters are still shared between NMT model and ILM, which might result in a better overall ILM approximation.\footnote{In their work, \citet{zeineldeen2021:ilm} used a mini-LSTM network with the same dependencies as our mini-self-attention.}
\end{description}

\section{Experiments}

\begin{figure}[ht]
\includegraphics[width=0.475\textwidth]{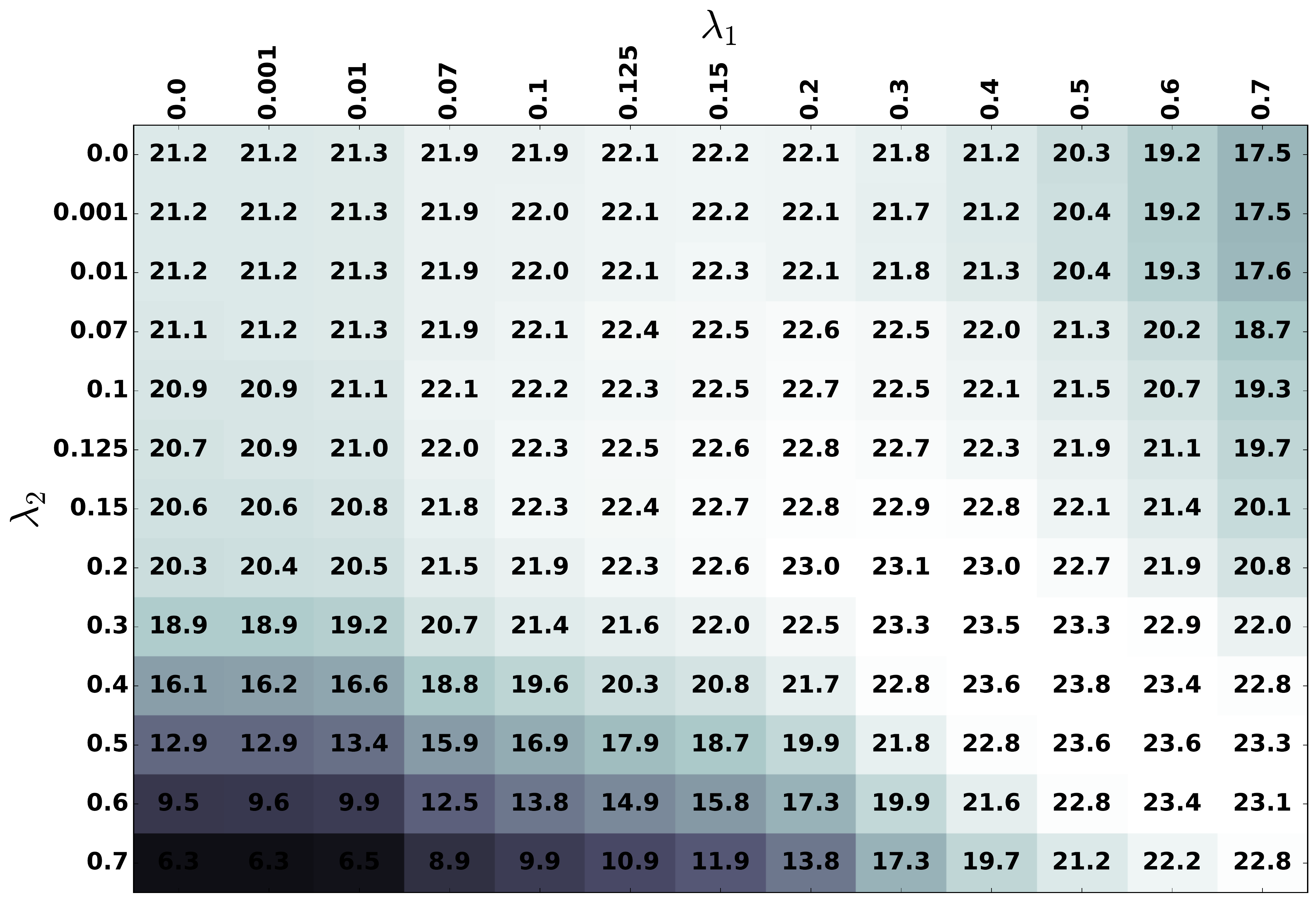}
  \caption{$\BLEU$ scores (percentage) on the validation set for the IWSLT En$\to$De task for different weights of LM and ILM (\textit{mini-self-attention}). $\lambda_1$ on the x-axis is the weight of the external LM while $\lambda_2$ on the y-axis is the (negative) weight of the ILM.}
\label{fig}
\end{figure}

We perform experiments on four machine translation tasks, representing different data conditions.
The exact data conditions and statistics are provided in the Appendix \ref{sec:appendix}.
For all tasks, the additional monolingual data, as well as the test sets, are in the news domain.
The monolingual data comes from Newscrawl\footnote{\url{https://data.statmt.org/news-crawl/}} where we sample ca. 10M sentences for LM training and back-translation.
For \textbf{IWSLT En$\to$De} and \textbf{IWSLT En$\to$It}, the parallel training data consists of around 200k sentence pairs and is in the scientific-talks-domain, coming from the IWSLT17 Multilingual Task \cite{cettolo2017overview}.
For this setting, we expect the biggest improvements from the additional monolingual data, since the parallel data is out-of-domain.
For \textbf{NEWS En$\to$De}, the parallel training data (around 300k sentence pairs) is in the news domain, coming from the NewsCommentaryV14 corpus\footnote{\url{https://data.statmt.org/news-commentary/v14/}}.
Finally, \textbf{WMT14 En$\to$De} is a standard NMT benchmark used by \citet{vaswani2017attention} where the parallel training data consists of around 3.9M sentence pairs and is of mixed domain.

We tokenize the data using byte-pair-encoding \cite{sennrich2016neural, DBLP:conf/acl/Kudo18} with 15k joint merge operations (40k for WMT14).
The models are implemented using the fairseq toolkit \cite{ott2019fairseq} following the transformer base architecture \cite{vaswani2017attention}.
The details of the training setups can be found in Appendix \ref{sec:appendix}.
All systems are trained until the validation perplexity no longer improves and the best checkpoint is selected using validation perplexity as well.
We use beam-search with beam-size 12 and utilize SacreBLEU \cite{post2018call} to calculate \BLEU \cite{papineni2002bleu} and \TER \cite{snover2006study}.
\begin{table}[t]
\centering
\begin{tabular}{l|r}
\toprule
Method              &  \multicolumn{1}{c}{valid-PPL} \\ \hline
\textit{separate LM}         &  109.9                        \\
$h=0$                 &  251.3                        \\
$h=h_{\text{avg}}$            &  240.9                        \\
$c=c_{\text{avg}}$            &  244.2                        \\
\textit{mini-self-attention} &  108.4                        \\ \bottomrule
\end{tabular}
\caption{Perplexities of the validation set for the IWSLT En-De task using different ILM model approximations.}
\label{tab:ppl_ILM}
\end{table}
\begin{table}[t]
\centering
\begin{tabular}{l|c|c|r|r}
\toprule
\multicolumn{1}{c|}{ILM} & $\lambda_1$ & $\lambda_2$ & \multicolumn{1}{c|}{\BLEU} & \multicolumn{1}{c}{\TER} \\ \hline
- & 0 & 0 & 28.9 & 52.8 \\ \hline
- & 0.15 & 0.0 & 30.0 & 52.3 \\
\textit{separate LM} & 0.5 & 0.3 & 31.2 & 50.9 \\
$h=0$ & 0.5 & 0.3 & 30.8 & 51.3 \\
$h=h_{\text{avg}}$ & 0.5 & 0.3 & 31.1 & 51.1 \\
$c=c_{\text{avg}}$ & 0.5 & 0.3 & 30.6 & 51.5 \\
\textit{mini-self-attn} & 0.5 & 0.4 & 31.7 & 50.0 \\ \bottomrule
\end{tabular}
\caption{Translation performance of the different ILM variants on the test set of the IWSLT En-De task. \BLEU and \TER are reported in percentage.}
\label{tab:IWSLT_en_de_final_results}
\end{table}
We report \BLEU and \TER since we are most familiar with these metrics and to be comparable with previous works.
However, we acknowledge that these metrics might have some biases and in future work it might be worth utilizing additional metrics like \COMET \cite{rei-etal-2020-comet} and \BLEURT \cite{sellam-etal-2020-bleurt}. 
Additionally, in future work we should separate our test sets for original source and target text to better understand the effect of translationese in both training and test data, as this might very much influence the improvements we see, especially in the case of back-translation \cite{freitag-etal-2020-bleu}.

\begin{table*}[t!]
\centering
\begin{tabular}{l|rr|rr|rr|rr}
\toprule
\multirow{2}{*}{Method} &
  \multicolumn{2}{c|}{IWSLT En-De} &
  \multicolumn{2}{c|}{IWSLT En-It} &
  \multicolumn{2}{c|}{NEWS En-De} &
  \multicolumn{2}{c}{WMT14 En-De} \\ \cline{2-9} 
 &
  \multicolumn{1}{c|}{\BLEU} &
  \multicolumn{1}{c|}{\TER} &
  \multicolumn{1}{c|}{\BLEU} &
  \multicolumn{1}{c|}{\TER} &
  \multicolumn{1}{c|}{\BLEU} &
  \multicolumn{1}{c|}{\TER} &
  \multicolumn{1}{c|}{\BLEU} &
  \multicolumn{1}{c}{\TER} \\ \hline
baseline \textit{external}   & \multicolumn{1}{r|}{-}    & -    & \multicolumn{1}{r|}{-}    & -    & \multicolumn{1}{r|}{$^{\dagger}$32.3}  & -    & \multicolumn{1}{r|}{$^{\ddagger}$27.3} & - \\ \hline
baseline \textit{ours}       & \multicolumn{1}{r|}{28.9} & 52.8 & \multicolumn{1}{r|}{24.1} & 58.9 & \multicolumn{1}{r|}{32.8} & 49.0 & \multicolumn{1}{r|}{27.7}        & 56.5  \\ 
\quad +SF      & \multicolumn{1}{r|}{30.0} & 52.3 & \multicolumn{1}{r|}{24.8} & 58.8 & \multicolumn{1}{r|}{33.2} & 49.8 & \multicolumn{1}{r|}{28.1}        & 56.6  \\
\quad \quad +ILM (\textit{separate LM})   & \multicolumn{1}{r|}{31.2} & 50.9 & \multicolumn{1}{r|}{26.0} & 57.8 & \multicolumn{1}{r|}{34.7} & 47.6  & \multicolumn{1}{r|}{28.8}        & 55.3  \\
\quad \quad +ILM (\textit{mini-self-attn}) & \multicolumn{1}{r|}{31.7} & 50.0 & \multicolumn{1}{r|}{26.1} & 57.0 & \multicolumn{1}{r|}{35.1} & 47.5  & \multicolumn{1}{r|}{29.1}        & 54.8  \\
back-translation                  & \multicolumn{1}{r|}{34.1} & 47.4 & \multicolumn{1}{r|}{27.2} & 56.9 & \multicolumn{1}{r|}{35.7} & 45.8  & \multicolumn{1}{r|}{29.5}        & 54.7  \\
\quad +SF +ILM (\textit{mini-self-attn})            & \multicolumn{1}{r|}{34.1} & 47.6 & \multicolumn{1}{r|}{27.3} & 56.7 & \multicolumn{1}{r|}{35.7} & 46.0  & \multicolumn{1}{r|}{29.8}        & 54.3  \\ \bottomrule
\end{tabular}
\caption{Comparison of LM fusion and back-translation on the four MT tasks. \BLEU and \TER are reported in percentage. External baselines are from $^{\dagger}$ \citet{kim2019and} and $^{\ddagger}$ \citet{vaswani2017attention}.}
\label{tab:final_results}
\end{table*}

\subsection{Comparison of ILM Approximations}
We start by analyzing the ILM neutralization approaches on the IWSLT En$\to$De task and then verify the results on the other tasks.

We implement and re-train (if applicable) all the different ILM approximation methods discussed in Section \ref{sec:methodology}.
The resulting perplexities on the validation set are listed in Table \ref{tab:ppl_ILM}.
The variants \textit{separate LM} and \textit{mini-self-attention} have been trained directly using the language model objective, so it is no surprise that they exhibit a much lower perplexity than the other approaches.
However, it can be argued that a lower perplexity of the ILM does not necessarily correspond to a better approximation of the implicit language model.

In order to effectively use the external LM and the ILM during decoding, we need to optimize the weights $\lambda_{1}$ and $\lambda_{2}$ (see Section \ref{sec:methodology}).
We do this via a grid search over the validation set by optimizing for the highest \BLEU score.
The resulting grid for the \textit{mini-self-attention} ILM variant on the IWSLT En$\to$De task is shown in Figure \ref{fig}. 

The NMT system by itself has a \BLEUpercent score of 21.2.
By log-linear combination with just the external LM ($\lambda_2=0$, vanilla shallow fusion) we can gain around 1\% absolute improvement on the validation set with the best choice of $\lambda_1 = 0.15$.
By including the ILM with a negative weight, we can get further improvements, up to a final score of 23.8 \BLEUpercent. \footnote{For the \textit{mini-self-attention} ILM variant, we also performed a more fine-grained search for $0.3 < \lambda_1, \lambda_2 < 0.6$ which did not result in further improvements.}
Interestingly, the best performance is reached when $\lambda_1 \approx \lambda_2$ and with the ILM neutralization, the external LM can be assigned a much bigger weight compared to the case $\lambda_2 = 0$.
We find that for all ILM approximation variants, the optimal weights are similar, and that the \TER scores on the validation set follow an almost identical pattern.
The final performance of each variant on the test set is shown in Table \ref{tab:IWSLT_en_de_final_results}.

We want to point out, that the improvements we see on the validation set transfer nicely to the test set with the same tuned weights $\lambda_1$ and $\lambda_2$.
This is because, in our experiments, the validation and test sets are of the exact same domain.
In some additional experiments we found that the optimal values for these weights are indeed domain specific and have to be re-tuned if the system were to be optimized for a different domain.
All ILM approximation variants lead to a significant performance improvement over simple shallow fusion.
Out of all ILM approximations, the \textit{mini-self-attention} approach performs best, which is the same observation that \citet{zeineldeen2021:ilm} made for ASR.

\subsection{Comparison to Back-Translation}

For the back-translation experiments, we train NMT systems on the same parallel training data in the reverse direction and then translate a total of 10M sentences from the monolingual target data (the same data used for training the external LM).
Afterwards, the final systems are trained on the combination of real and synthetic data.
The final results for all four MT tasks are shown in Table \ref{tab:final_results}.

We observe the same trend for all four MT tasks.
In general, the improvements from the additional monolingual data are getting smaller, when the amount of parallel training data increases. 
In almost all cases, shallow fusion gives a small improvement over just using the NMT system.
ILM neutralization again improves consistently over simple shallow fusion, with the \textit{mini-self-attn} approximation variant always performing the best.
Back-translation out-performs language model integration on all four tasks, although the gap is getting smaller the more parallel training data is available.

We also combine back-translation with the best ILM approximation approach (\textit{mini-self-attn}).
This does not further increase translation quality, with the exception of the WMT14 task, where we see a small improvement.
In general, the ILM approach performs the closest to back-translation on the WMT14 task, so it might be worthwhile to apply this concept to an even bigger MT task.

\section{Conclusion}

We re-visit the method of language model integration for neural machine translation.
We implement and experiment with a new approach of neutralizing the implicit language model, which has already shown promising result for the task of automatic speech recognition.
We find that ILM neutralization significantly improves the translation quality compared to standard shallow fusion.
However, back-translation as an alternative way to incorporate additional monolingual data, still outperforms the approaches using an external language model.
Therefore, for future work we will focus on scenarios where back-translation can not be applied effectively, e.g. when the quality of the initial NMT system is too bad to create helpful synthetic data.

\section*{Acknowledgements}
This work was partially supported by the project HYKIST funded by the German Federal Ministry of Health on the basis of a decision of the German Federal Parliament (Bundestag) under funding ID ZMVI1-2520DAT04A, and by NeuroSys which, as part of the initiative “Clusters4Future”, is funded by the Federal Ministry of Education and Research BMBF (03ZU1106DA).

\section*{Limitations}

The approach of language model integration for neural machine translation is analyzed and compared to the de-facto standard method of back-translation.
Due to constrained resources, this work has several limitations.
We focus on translation of text in a single domain, namely news-articles.
Different domains might exhibit different behaviour.
For the back-translation experiments, we use beam search to create the synthetic data, other methods like sampling were not considered.
When combining the synthetic and real parallel data, there are additional methods like tagging and block-wise batching, which we did not utilize in this work.
Finally, we compare against the most commonly used LM fusion approach, i.e. shallow fusion.
There exist other LM fusion techniques which might exhibit different behaviour when used in combination with ILM neutralization.


\bibliography{anthology,custom}
\bibliographystyle{acl_natbib}

\clearpage

\appendix

\section{Appendix}
\label{sec:appendix}

All validation and test sets are from the WMT news translation tasks \cite{farhad2021findings}.
The validation/test sets are WMT newstest2015/newstest2018 for IWSLT En$\to$De and NEWS En$\to$De, newssyscomb2009/newstest2009 for IWSLT En$\to$It and newstest2013/newstest2014 for WMT14 En$\to$De.
Data statistics can be found in Table \ref{tab:data}.

\begin{table}[h!]
\centering
\begin{tabular}{l|r|r|r}
\hline
\multicolumn{1}{c|}{task} & \multicolumn{1}{c|}{dataset} & \multicolumn{1}{c|}{domain} & \multicolumn{1}{c}{\# sent.} \\ \hline
IWSLT & train & scientific-talks & 210k \\
En$\to$De & valid & news & 2.2k \\
 & test & news & 3k \\
 & mono. & news & 9.7M \\ \hline
IWSLT & train & scientific-talks & 232k \\
En$\to$It & valid & news & 500 \\
 & test & news & 2.5k \\
 & mono. & news & 10.0M \\ \hline
NEWS & train & news & 330k \\
En$\to$De & valid & news & 2.2k \\
 & test & news & 3k \\
 & mono. & news & 9.7M \\ \hline
WMT14 & train & mixed & 3.9M \\
En$\to$De & valid & news & 3k \\
 & test & news & 3k \\
 & mono. & news & 10.0M \\ \hline
\end{tabular}
\caption{Data statistics for all tasks.}
\label{tab:data}
\end{table}

We use dropout 0.3 and label-smoothing 0.2 for IWSLT En$\to$De, IWSLT En$\to$It and NEWS En$\to$De and dropout 0.3 and label-smoothing 0.1 for WMT14 En$\to$De.
The resulting NMT models had ca. 51M parameters for IWSLT En$\to$De, IWSLT En$\to$It and NEWS En$\to$De and ca. 67M parameters for WMT14 En$\to$De.
The NMT training took around 24h for IWSLT En$\to$De, IWSLT En$\to$It and NEWS En$\to$De and around 150h for WMT14 En$\to$De on a single NVIDIA GeForce RTX 2080 Ti graphics card.
The language models had ca. 26M parameters for IWSLT En$\to$De, IWSLT En$\to$It and NEWS En$\to$De and ca. 41M parameters for WMT14 En$\to$De.
All language model trainings took around 150h on a single NVIDIA GeForce RTX 2080 Ti graphics card.
Due to computational limitations, we report results only for a single run.

\end{document}